\pdfoutput=1
\PassOptionsToPackage{dvipsnames}{xcolor}
\documentclass[11pt]{article}

\usepackage[final]{acl}

\usepackage{times}
\usepackage{latexsym}
\usepackage{amsmath,amssymb}

\usepackage{tabularx}
\usepackage{multirow}
\usepackage{booktabs}
\usepackage{adjustbox}
\usepackage{caption}
\usepackage{subcaption}
\usepackage{makecell} 

\usepackage{pifont}
\newcommand{\cmark}{\ding{51}}%
\newcommand{\xmark}{\ding{55}}%
\usepackage[dvipsnames]{xcolor}

\usepackage[T1]{fontenc}

\usepackage[utf8]{inputenc}

\usepackage{microtype}

\usepackage{inconsolata}

\usepackage{graphicx}

\title{Composable Cross-prompt Essay Scoring by Merging Models}

\author{Sanwoo Lee,
Kun Liang,
Yunfang Wu\thanks{~~Corresponding author.} \\
National Key Laboratory for Multimedia Information Processing, Peking University \\ 
School of Computer Science, Peking University \\
\texttt{\{sanwoo, wuyf\}@pku.edu.cn}}

\begin{document}
\maketitle
\begin{abstract}
Recent advances in cross-prompt automated essay scoring (AES) typically train models jointly on all source prompts, often requiring additional access to unlabeled target prompt essays simultaneously. However, using all sources is suboptimal in our pilot study, and re-accessing source datasets during adaptation raises privacy concerns. We propose a \textit{source-free} adaptation approach that selectively \textit{merges} individually trained source models' parameters instead of datasets. In particular, we simulate joint training through linear combinations of task vectors---the parameter updates from fine-tuning. To optimize the combination's coefficients, we propose \textbf{P}rior-encoded \textbf{I}nformation \textbf{M}aximization (\textbf{PIM}), an unsupervised objective which promotes the model's score discriminability regularized by priors pre-computed from the sources. We employ Bayesian optimization as an efficient optimizer of PIM. Experimental results with LLMs on in-dataset and cross-dataset adaptation show that our method (1) consistently outperforms training jointly on all sources, (2) maintains superior robustness compared to other merging methods, (3) excels under severe distribution shifts where recent leading cross-prompt methods struggle, all while retaining computational efficiency.

\end{abstract}

\section{Introduction}

Automated essay scoring (AES) is a machine learning task of developing a system that scores essays written in response to a given prompt (i.e., writing instructions). A prompt represents a domain as different prompts may have distinct topics. Early works had achieved success in prompt-specific AES \citep{chen-he-2013-automated, taghipour-ng-2016-neural, dong-etal-2017-attention, farag-etal-2018-neural} where test samples were assumed to belong to the same prompt as training samples. Yet prompt-specific models were found to struggle when tested on new prompts \citep{phandi2015Flexible}, accelerating efforts on cross-prompt AES with domain adaptation or generalization techniques \citep{zesch-etal-2015-task, jin-etal-2018-tdnn, chen-li-2023-pmaes, jiang-etal-2023-improving}. 

Despite the adaptability to the data-scarce target prompt, current cross-prompt methods typically assume simultaneous access to the data from both the source and target prompts if they leverage target samples for adaptation \citep{cao2020domain, chen-li-2023-pmaes}. Yet this assumption is often violated due to privacy concerns in disclosing essays. Instead, models trained from the source prompts are safer to distribute. Hence adapting without source datasets holds great practical implications, which aligns with the unsupervised source-free domain adaptation (SFDA) paradigm \citep{pmlr-v119-liang20a, wang2021tent, HCL-21}.

\begin{figure}[t]
    \centering
    \includegraphics[width=\columnwidth]{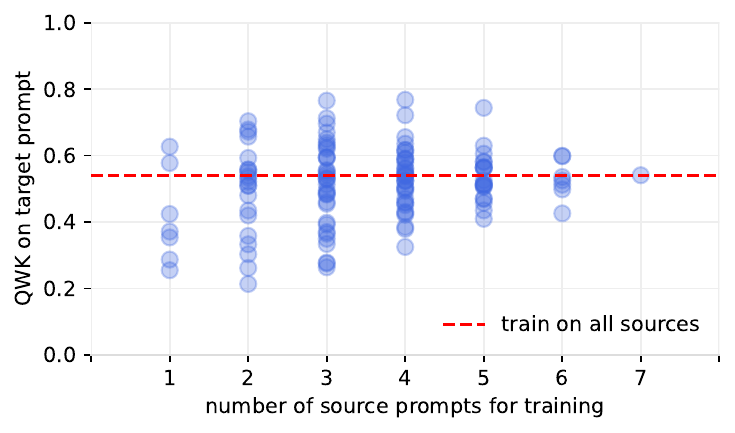}
    \caption{Agreement with human raters (QWK) on target prompt (P7 of ASAP), using BERT \citep{devlin-etal-2019-bert} fine-tuned jointly on varying number of source prompt datasets. Notably, training on all sources leads to suboptimal performance. Similar trends are observed on other target prompts, see Appendix~\ref{apdx:pilot_study_full}.}
    \label{fig:pilot_study}
\end{figure}

On the other hand, selecting the most relevant source domains remains a crucial yet underexplored aspect in cross-prompt AES. Most works either adopt single-source adaptation setting \citep{phandi2015Flexible, dong-zhang-2016-automatic, cozma-etal-2018-automated}, or train the model jointly on all source domain datasets for multi-source adaptation \citep{jin-etal-2018-tdnn, ridley2021automated, do-etal-2023-prompt, chen-li-2024-plaes} likely due to the conventional belief that more sources yield better performance. However, our pilot study in Figure \ref{fig:pilot_study} shows that carefully selecting a subset of source prompt datasets clearly outperforms training on all sources. Nevertheless, the high cost of joint training makes exhaustive search for the optimal subset impractical.

In this work, we explore \textit{merging models} \citep{wortsman2022model, matena2022merging, ainsworth2023git} as a scalable alternative to joint training for source-free domain adaptation in cross-prompt AES. That is, we combine models fine-tuned on individual source prompts without re-training. In particular, the weighted sum of the models' task vectors \citep{ilharco2023editing}---parameter updates after fine-tuning---is added back to the pre-trained model (Eq.~\ref{eq:task_arithmetic}), which effectively mimics joint training in a post-hoc fashion. It then allows for fast and iterative search over the mixing coefficients that (soft-) select the task vectors. 

To guide this search in the absence of the target labels and source datasets, we propose \textbf{P}rior-encoded \textbf{I}nformation \textbf{M}aximization (\textbf{PIM}), an information-theoretic objective that leverages useful priors pre-computed from the labeled source domains, to enhance scoring performance (Sec.~\ref{sec:objective_function}). The objective is coupled with Bayesian optimization for an efficient optimization of PIM (Sec.~\ref{sec:bayes_opt}). 

We merge lightweight LoRA adapters \citep{hu2022lora} of large language models (LLMs), motivated by their extensibility to generate rationales \citep{chu-etal-2025-rationale} and the surging efforts to build ever-stronger LLMs. Experiments on in- and cross-dataset settings show that: our method (1) outperforms training jointly on all sources, (2) outperforms other merging strategies in numerous cases, (3) remains robust under severe shifts (i.e., cross-dataset) while recent cross-prompt AES methods underperform, and (4) is  time-saving than adaptation method training jointly on all sources.

In summary, our contributions are as follows:
\begin{itemize}
    \item We propose a domain-adaptive model merging approach for source-free cross-prompt AES. See Table~\ref{tab:comparison_of_settings} for the comparison of settings.
    \item We design an unsupervised objective which promotes model's score discriminability regularized by priors derived from the sources.   
    \item Beyond in-dataset, we validate our method in cross-dataset adaptation, under which our method remains more robust than recent cross-prompt  methods.
\end{itemize}

\begin{table}[t]
    \centering
    \small
    \resizebox{\columnwidth}{!}{
    \begin{tabular}{ccccc}
        \toprule
        \textbf{Method} & \textbf{\makecell{multi-\\source}} & \textbf{\makecell{unlabeled\\target data}} & \textbf{\makecell{free of\\source data}} & \textbf{\makecell{source \\ selection}} \\
        \midrule
        \citet{phandi2015Flexible} & \textcolor{Red}{\xmark} & \textcolor{Red}{\xmark} & \textcolor{Red}{\xmark} & \textcolor{Red}{\xmark} \\ 
        \citet{cao2020domain} & \textcolor{Red}{\xmark} & \textcolor{OliveGreen}{\cmark} & \textcolor{Red}{\xmark} & \textcolor{Red}{\xmark} \\
        \citet{ridley2020prompt} & \textcolor{OliveGreen}{\cmark} & \textcolor{Red}{\xmark} & \textcolor{OliveGreen}{\cmark} & \textcolor{Red}{\xmark} \\
        \citet{chen-li-2023-pmaes} & \textcolor{OliveGreen}{\cmark} & \textcolor{OliveGreen}{\cmark} & \textcolor{Red}{\xmark} & \textcolor{Red}{\xmark} \\
        Ours & \textcolor{OliveGreen}{\cmark} & \textcolor{OliveGreen}{\cmark} & \textcolor{OliveGreen}{\cmark} & \textcolor{OliveGreen}{\cmark} \\ 
        \bottomrule
    \end{tabular}}
    \caption{Comparison of adaptation settings among holistic cross-prompt AES methods based on key criteria. Our proposed setting satisfies all listed criteria.}
    \label{tab:comparison_of_settings}
\end{table}

\section{Preliminary}

\begin{figure*}[t]
    \centering
    \includegraphics[width=1.0\textwidth]{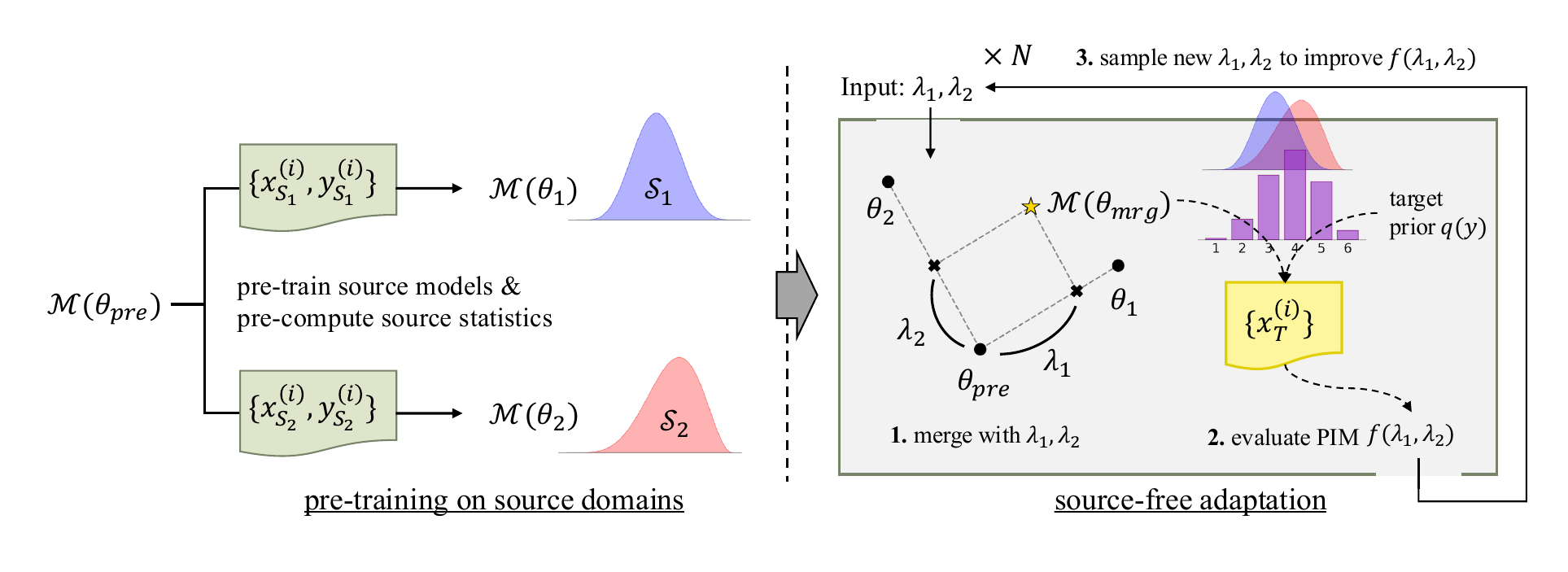}
    \caption{An illustration of our method for source-free cross-prompt AES. \textbf{Left}: Source models and statistics are pre-trained before adaptation. \textbf{Right}: During source-free adaptation, merging coefficients are optimized via Bayesian optimization to enhance our prior-encoded information maximization (PIM) criterion (Eq. \ref{eq:objective}).}
    \label{fig:overview}
    \vspace{-3mm}
\end{figure*}
\paragraph{Problem Statement.} 
This paper uses "prompt" and "domain" interchangeably. We consider the unsupervised source-free domain adaptation problem \citep{pmlr-v119-liang20a, wang2021tent, yang2022attracting} with multiple source domains for cross-prompt AES, where the input $x \in \mathcal{V}^*$ is a sequence of tokens and the output $y \in \mathbb{Z}$ is an integer score. A pre-trained model $\mathcal{M}(\theta_{pre})$ is fine-tuned on each one of the source datasets separately, and additional statistics $\mathcal{S}_j$ may be computed for each source, after which the source datasets become no longer available \citep{adachi2025testtime}. We denote $\mathcal{M}(\theta_{j})$ as the model fine-tuned on $j$-the source dataset $\mathcal{D}_{S_j} =\{(x_{S_j}^{(i)}, y_{S_j}^{(i)})\}_{i=1}^{N_{j}}$, parameterized by $\theta_j \in \mathbb{R}^d$. Note that separating each source domain apart is not a requirement for source-free adaptation, but is a stricter setting we aim to address via model merging.

During the adaptation phase, we have access to the set of fine-tuned models $\{f(\theta_j)\}_{j=1}^{M}$, an unlabeled target dataset $\mathcal{D}_T = \{(x_T^{(i)})\}_{i=1}^{N_T}$ and optionally, the summary statistics $\{\mathcal{S}_j\}_{j=1}^{M}$. As a side note, the score range may vary across domains (e.g., Table~\ref{tab:dataset_stats}). During inference, we require the model to adapt to potentially novel score ranges, different from the common approach that normalizes all score ranges to a shared scale \citep{taghipour-ng-2016-neural, cozma-etal-2018-automated, wang-liu-2025-mes}.

\paragraph{LLM Regression.} 
We employ LLMs for scoring essays—a regression task implemented by autoregressive generation. Following \citet{lukasik2025better}, we assume each score $y$ corresponds to a unique string representation $\text{str}(y) \in \mathcal{V}^*$ (e.g., $2 \rightarrow \text{"2"} $). Each input-output pair $(x, y)$ is transformed into a formatted pair $(x', y')$ using an instruction template (e.g., see Appendix~\ref{apdx:instruction_template}). The pre-trained model $\mathcal{M}(\theta_{pre})$ is then instruction-tuned on the source dataset $\mathcal{D}_j$ to maximize the likelihood of generating the answer:
\begin{equation}
    \theta_j = \mathop{\arg \max}_{\theta}\mathop{\mathbb{E}}_{x,y} \left[ \log p(y'|x', \theta) \right]
\end{equation}

\paragraph{Model Merging.}

The seminal work of \citet{ilharco2023editing} introduced the concept of a \textit{task vector} $\tau_j:=\theta_j-\theta_{pre}$ defined as the parameter updates obtained through fine-tuning. Interestingly, adding task vectors from multiple tasks to the pre-trained model has been shown to effectively approach multi-task training albeit with a performance gap, a finding further verified by follow-up studies \citep{yadav-ties, yu2024language, deep2024DELLAMerging}. We follow this merging framework and model the source selection and joint training through a linear combination of task vectors:
\begin{equation}
\label{eq:task_arithmetic}
    \theta_{mrg} = \theta_{pre} + \sum_{j=1}^M  \lambda_i \tau_j
\end{equation}
where $\{\lambda_j \in \mathbb{R}\}_{j=1}^{M}$ are the mixing coefficients. 

Updating entire parameters results in large task vectors, making merging inefficient. We instead adopt low-rank adaptation (LoRA) \citep{hu2022lora} and merge lightweight adapters. LoRA fine-tunes models by learning low-rank updates: for a weight matrix $W \in \mathbb{R}^{m \times n}$, the update is $W + \Delta W = W + BA$, where $B \in \mathbb{R}^{m \times r}$ and $ A \in \mathbb{R}^{r \times n}$ are learnable matrices ($r \ll \min(m, n)$). Accordingly, we define task vectors in terms of LoRA adapters:

\begin{equation}
\label{eq:lora_task_vector}
    \tau_j = \theta_j - \theta_{pre}= \underset{l=1} {\overset{L} {\Large||}}  \text{flatten}(B_j^{(l)} A_j^{(l)})
\end{equation}
where $\Large||$ denotes concatenation of the vectors $ \text{flatten}(B_j^{(l)} A_j^{(l)})$ across $L$ layers. Layers without adapters contribute zeros. In practice, these LoRA-induced task vectors can be computed efficiently without explicit multiplication $B_j^{(l)} A_j^{(l)}$ and zero-padding to achieve Eq.~\ref{eq:task_arithmetic}.

Given this setup, merging models via a linear combination of LoRA adapters (Eq.~\ref{eq:task_arithmetic}, \ref{eq:lora_task_vector}) reduces our objective to selecting coefficients $\lambda_1, \dots, \lambda_M$ that leads to an  optimal performance on the target prompt.

\section{Method}
Given the absence of target labels in cross-prompt essay scoring, we establish an objective that promotes the model's scoring performance from an information-theoretic view (Sec.~\ref{sec:objective_function}), and employ Bayesian optimization to maximize this objective without backpropagation (Sec.~\ref{sec:bayes_opt}). Figure \ref{fig:overview} illustrates an overview of our method.

\subsection{Prior-encoded Information Maximization}
\label{sec:objective_function}

In essay scoring where target labels are ordinal scores, a plausible scoring model would assign unambiguous labels for individual essays, while retaining the discriminability across different essays. In standard classification setup, this idea has been formalized as maximizing the mutual information (\textbf{MI}) between the input $x$ and the output $y$ \citep{unsupcls91bridle, discluster10krause, pmlr-v119-liang20a}. In what follows, we revisit this principle, study which of its properties can be modified to be better applied in essay scoring, and propose our final objective.

The MI between input $x$ and output $y$ under a discriminative model $p(y|x, \theta)$ is given by:
\begin{equation}
    \mathcal{I}(y;x) = \mathcal{H}(p(y|\theta)) - \mathcal{H}(p(y|x, \theta))
\end{equation}
where $\mathcal{H}(\cdot)$ denotes entropy. In classification, $\mathcal{I}(y;x)$ is empirically estimated as
\begin{align}
    \mathcal{I}(y;x) &=  \mathcal{H}(p(y|\theta)) - \frac{1}{N} \sum_{i=1}^{N} \mathcal{H}(p(y|x^{(i)}, \theta)), \\
    p(y|\theta) &= \frac{1}{N} \sum_{i=1}^{N} p(y|x^{(i)}, \theta)
\end{align}
in which $p(y|x^{(i)}, \theta) \in \mathbb{R}^C$ denotes the output probability of a sample $x^{(i)}$. Essentially, maximizing $\mathcal{I}(y;x)$ balances between sample-wise sharpness and global discriminability in predictions. 

However, directly applying this criterion to (ordinal) essay scoring can be problematic. Note that maximizing $\mathcal{H}(p(y|\theta)) = \log C - KL(p(y|\theta)||U)$ is equivalent to minimizing the KL-divergence between the marginal distribution $p(y|\theta)$ and a uniform distribution $U$. Given the classes are sorted discrete scores in our setting, $U$ is unlikely to be the true marginal $p(y)$, since assigning extreme scores are less likely than the mid-range ones.  

Based on this insight, we propose \textbf{P}rior-encoded \textbf{I}nformation \textbf{M}aximization (\textbf{PIM}), where we extract an informative prior $q(y)$ from the labeled source domains, and use it in place of $U$ for MI maximization. To suit source-free adaptation, we pre-compute the marginal distribution $p(y)$ for each source domain \textit{before} removing the dataset. In particular, for $j$-th source domain, we first scale the scores $\{  y_{S_j}^{(i)} \}_{i=1}^{N_{j}}$ to the $[0, 1]$ interval:
\begin{equation}
    \tilde{y}_{S_j}^{(i)} = (y_{S_j}^{(i)} - a_j+0.5)/(b_j - a_j+1)~~~\forall i
\end{equation}
where $y_{S_j}$ is assumed to be an integer ranging from $a_j$ to $b_j$. Next, we fit a Beta distribution $\mathbf{Beta}(\alpha_j, \beta_j)$ with the scaled scores via maximum likelihood estimation:
\begin{equation}
    (\alpha_j, \beta_j) = \mathop{\arg\max}_{(\alpha_j, \beta_j)}~~ \mathbb{E}_{\tilde{y}_j}\left[ \log \mathbf{Beta}(\tilde{y}_j;\alpha_j, \beta_j) \right]
\end{equation}
where $\mathbf{Beta}(\tilde{y}_j;\alpha_j, \beta_j)$ is the probability density at $\tilde{y}_j$. $\mathbf{Beta}(\alpha, \beta)$ is a suitable abstraction of a set of noisy (scaled) scores, given that it is unimodal when $\alpha>1, \beta>1$ and flexible in modeling the skewness/dispersion of the distribution bounded by $[0, 1]$, just as essay scores being bounded and roughly unimodal.

During the \textit{adaptation} stage, we unify all source Beta distributions into a single $\mathbf{Beta}(\alpha_{\mathcal{S}}, \beta_{\mathcal{S}})$ to further reduce domain-specific noise. Essentially, we consider the mean $\mu$ and variance $\sigma^2$ of the mixture $1/M\sum_{j=1}^{M} \mathbf{Beta}(\alpha_j, \beta_j)$ and set $\mathbf{Beta}(\alpha_{\mathcal{S}}, \beta_{\mathcal{S}})$ such that its mean and variance equal to $\mu$ and $\sigma^2$ (derivations in Appendix \ref{apdx:moment_matching}). This unified distribution is then discretized into a categorical distribution $q(y) \in \mathbb{R}^{C_{T}}$ over the target prompt (sorted) score classes: 
\begin{equation}
     q_c(y) = \int_{\frac{c-1}{C_T}}^{\frac{c}{C_T}} \mathbf{Beta}(y; \alpha_S, \beta_S)\, dy, \text{for } c \in 1{:}C_T
\end{equation}
yielding source-informed prior probabilities over $C_T$ evenly spaced bins.

Finally, we define our PIM objective $f(\lambda)$ as maximizing the prior-encoded mutual information by inserting $q(y)$ in place of $U$:
\begin{equation}
\label{eq:objective}
    f(\lambda) = -KL(p(y|\lambda)||q(y)) -\mathcal{H}(p(y|x , \lambda))
\end{equation}
Here, the objective is written in terms of the merging coefficients $\lambda=[\lambda_1, \dots, \lambda_M]^\top \in \mathbb{R}^{M}$ to explicitly state that the parameter $\theta$ is solely determined by $\lambda$ in our chosen merging framework (Eq. \ref{eq:task_arithmetic}). In the context of using LLMs, we define $p(y|x^{(i)}, \lambda) \in \mathbb{R}^{C_T}$ to be the next token probabilities of the token that predicts the score, where the probabilities are further truncated to $C_T$ score-representing tokens and subsequently normalized to sum to $1$.\footnote{For our instruction template (Appendix \ref{apdx:instruction_template}), $p(y|x^{(i)}, \lambda)$ is the post-processed next-token probabilities of the special token that starts the assistant’s response (e.g., <assistant>).}

\subsection{Bayesian Optimization}
\label{sec:bayes_opt}

In determining the mixing coefficients $\lambda$, recent studies have shown success in using Bayesian optimization \citep{nips24bayesianfusion, liu2024checkpoint} which is computationally less demanding than training the coefficients \citep{wortsman2022model}. Following this approach, we leverage Bayesian optimization to maximize $f(\lambda)$ (Eq. \ref{eq:objective}) in terms of $\lambda \in \mathbb{R}^{M}$ without backpropagation. 

Essentially, the algorithm treats $f(\lambda)$ as a black-box function and constructs a surrogate of $f(\lambda)$ as a sample from a Gaussian Process---a distribution over functions \citep{williams2006gaussian}. Given the prior mean function $\mu_0$, covariance function $\Sigma_0$ and $k$ observations $f(\lambda^{(1:k)}):=\{f(\lambda^{(i)})\}_{i=1}^{k}$, it updates the \textit{posterior} distribution over the function value at the current $(k+1)$-th iteration, i.e., $f(\lambda^{(k+1)})|f(\lambda^{(1:k)})$. Next, the \textit{acquisition function} determines where to sample $\lambda^{(k+1)}$ based on the posterior. In particular, we use Expected Improvement (\textbf{EI}) which maximizes the expected gain over the current best value $f^*(k):=\mathop{\max}_{\lambda^{(i)}}\{f(\lambda^{(i)})\}_{i=1}^{k}$:
\begin{equation}
    \mathop{\arg \max}_{\lambda^{(k+1)}} \mathop{\mathbb{E}}_{f(\lambda^{(k+1)})}\left[ \max(f(\lambda^{(k+1)}) - f^*(k), 0) \right]
\end{equation}
This process of posterior estimation and next point sampling is repeated until convergence. The final solution is $\mathop{\arg\max}_{\lambda^{(i)}}\{f(\lambda^{(i)})\}_{i=1}^{N}$ for $N$ total iterations. See Appendix~\ref{apdx:bayes_opt_details} for additional details.

\section{Experiment}

\begin{table}[t]
    \centering
    \small
    \resizebox{\columnwidth}{!}{
    \begin{tabular}{cccccc}
        \toprule
        \textbf{Dataset} & \textbf{Prompt} & \textbf{\#Essay} & \textbf{Genre} & \textbf{Avg Len} & \textbf{Range} \\
        \midrule
        \multirow{8}{*}{\textbf{ASAP}}
         & 1 & 1783 & ARG & 427 & 2-12 \\
         & 2 & 1800 & ARG & 432 & 1-6 \\
         & 3 & 1726 & RES & 124 & 0-3 \\
         & 4 & 1772 & RES & 106 & 0-3 \\
         & 5 & 1805 & RES & 142 & 0-4 \\
         & 6 & 1800 & RES & 173 & 0-4 \\
         & 7 & 1569 & NAR & 206 & 0-30 \\
         & 8 & 723 & NAR & 725 & 0-60 \\
        \midrule
        \multirow{8}{*}{\textbf{PERSUADE2.0}}
        & 1 & 1656 & ARG & 339 & 1-6 \\
         & 2 & 2157 & ARG & 641 & 1-6 \\
         & 3 & 1670 & ARG & 552 & 1-6 \\
         & 4 & 1552 & ARG & 573 & 1-6 \\
         & 5 & 1372 & RES & 330 & 1-6 \\
         & 6 & 2046 & RES & 455 & 1-6 \\
         & 7 & 1862 & RES & 399 & 1-6 \\
         & 8 & 1583 & RES & 381 & 1-6 \\
        \bottomrule
    \end{tabular}}
    \caption{Dataset Statistics. \textbf{Genre}: ARG (argumentative), RES (source-dependent), NAR (narrative). \textbf{Avg Len}: Average essay length in words. \textbf{Range}: Score range.}
    \label{tab:dataset_stats}
\end{table}

\begin{table*}[t]
    \small
    \centering
    \begin{adjustbox}{width=\textwidth,center}
    \begin{tabular}{cclccccccccc}
    \toprule
    \textbf{Model} & \textbf{Scheme} & \textbf{Method} 
    & \textbf{P1} & \textbf{P2} & \textbf{P3} & \textbf{P4} & \textbf{P5} & \textbf{P6} & \textbf{P7} & \textbf{P8} & \textbf{Avg.} \\
    \midrule
    gpt-4.1-mini & zero-shot & - & 0.063 & 0.423 & 0.459 & 0.672 & 0.480 & 0.624 & 0.321 & 0.390 & 0.429  \\
    \cmidrule{1-12}
    \multirow{10.25}{*}{llama-3.1-8b-it}
    & zero-shot & - & 0.109 & 0.246 & 0.239 & 0.240 & 0.361 & 0.407 & 0.321 & 0.484 & 0.301 \\
    \cmidrule{2-12}
    & \multirow{7}{*}{merge}
    & Averaging & 0.526 & 0.465 & 0.527 & 0.593 & 0.720 & 0.738 & 0.608 & 0.163 & 0.542* \\
    & & Fisher Merging & 0.437 & 0.541 & 0.521 & 0.590 & 0.670 & 0.724 & 0.562 & 0.167 & 0.526* \\
    & & RegMean & 0.482 & 0.461 & 0.526 & 0.580 & 0.724 & 0.731 & 0.580 & 0.135 & 0.527* \\
    & & Task Arithmetic & 0.787 & 0.368 & 0.604 & 0.632 & 0.772 & 0.741 & 0.627 & 0.120 & 0.581* \\
    & & TIES-Merging & 0.582 & 0.527 & 0.532 & 0.619 & 0.711 & 0.752 & 0.595 & 0.155 & 0.559* \\
    & & AdaMerging & 0.756 & 0.285 & 0.577 & 0.619 & 0.767 & 0.664 & 0.661 & 0.059 & 0.548* \\
    & & PIM (Ours) & 0.682 & 0.562 & 0.612 & 0.647 & 0.762 & 0.690 & 0.711 & 0.152 & \bf0.602 \\
    \cmidrule{2-12}
    & joint-train & - & 0.606 & 0.512 & 0.611 & 0.656 & 0.743 & 0.760 & 0.666 & 0.257 & 0.601 \\
    \cmidrule{1-12}
    \multirow{10.25}{*}{phi-4-mini-it}
    & zero-shot & - & 0.084 & 0.305 & 0.238 & 0.479 & 0.367 & 0.350 & 0.131 & 0.184 & 0.267 \\
    \cmidrule{2-12}
    & \multirow{7}{*}{merge}
    & Averaging & 0.383 & 0.613 & 0.494 & 0.625 & 0.528 & 0.652 & 0.396 & 0.334 & 0.503* \\
    & & Fisher Merging & 0.348 & 0.625 & 0.490 & 0.617 & 0.503 & 0.637 & 0.389 & 0.299 & 0.489* \\
    & & RegMean & 0.348 & 0.607 & 0.507 & 0.619 & 0.577 & 0.666 & 0.378 & 0.286 & 0.498* \\
    & & Task Arithmetic & 0.772 & 0.334 & 0.618 & 0.654 & 0.690 & 0.684 & 0.683 & 0.211 & 0.581* \\
    & & TIES-Merging & 0.532 & 0.568 & 0.512 & 0.625 & 0.542 & 0.681 & 0.448 & 0.291 & 0.525* \\
    & & AdaMerging & 0.742 & 0.316 & 0.569 & 0.618 & 0.645 & 0.650 & 0.608 & 0.227 & 0.547* \\
    & & PIM (Ours) & 0.737 & 0.585 & 0.637 & 0.612 & 0.731 & 0.654 & 0.692 & 0.387 & \bf0.629\\
    \cmidrule{2-12}
    & joint-train & - & 0.578 & 0.469 & 0.622 & 0.655 & 0.668 & 0.740 & 0.590 & 0.376 & 0.587* \\
    \bottomrule
    \end{tabular}
    \end{adjustbox}
    \caption{\textbf{In-dataset} cross-prompt evaluation results on \textbf{ASAP $\rightarrow$ ASAP}, measured by QWK. \textbf{P1-8} denotes Prompt 1-8. For each held-out target prompt, other $7$ prompts constitute the source domains. *: Significant improvement ($p<0.05$) of our method over a merging baseline/joint-training in Avg. QWK. The best average QWK is boldfaced.}
    \label{tb:in-dataset_result}
\end{table*}

\begin{table*}[ht]
    \small
    \centering
    \begin{adjustbox}{width=\textwidth,center}
    \begin{tabular}{cclccccccccc}
    \toprule
    \textbf{Model} & \textbf{Scheme} & \textbf{Method} 
    & \textbf{P1} & \textbf{P2} & \textbf{P3} & \textbf{P4} & \textbf{P5} & \textbf{P6} & \textbf{P7} & \textbf{P8} & \textbf{Avg.} \\
    \midrule
    \multirow{10.25}{*}{llama-3.1-8b-it}
    & zero-shot & - & 0.136 & 0.363 & 0.278 & 0.309 & 0.043 & 0.120 & 0.166 & 0.139 & 0.194\\
    \cmidrule{2-12}
    & \multirow{7}{*}{merge}
    & Averaging & 0.365 & 0.529 & 0.407 & 0.397 & 0.222 & 0.463 & 0.342 & 0.296 & 0.378*\\
    & & Fisher Merging & 0.420 & 0.599 & 0.472 & 0.479 & 0.248 & 0.497 & 0.392 & 0.348 & 0.432*\\
    & & RegMean & 0.340 & 0.513 & 0.391 & 0.399 & 0.208 & 0.441 & 0.322 & 0.291 & 0.363*\\
    & & Task Arithmetic & 0.412 & 0.377 & 0.310 & 0.311 & 0.164 & 0.294 & 0.264 & 0.340 & 0.309*\\
    & & TIES-Merging & 0.474 & 0.551 & 0.438 & 0.427 & 0.275 & 0.511 & 0.368 & 0.343 & 0.423*\\
    & & AdaMerging & 0.396 & 0.122 & 0.116 & 0.151 & 0.153 & 0.216 & 0.206 & 0.270 & 0.204*\\
    & & PIM (Ours) & 0.504 & 0.734 & 0.674 & 0.652 & 0.197 & 0.464 & 0.420 & 0.449 & \bf0.512\\
    \cmidrule{2-12}
    & joint-train & - & 0.515 & 0.406 & 0.438 & 0.448 & 0.243 & 0.367 & 0.342 & 0.454 & 0.401*\\
    \cmidrule{1-12}
    \multirow{10.25}{*}{phi-4-mini-it}
    & zero-shot & - & 0.181 & 0.216 & 0.397 & 0.515 & 0.341 & 0.345 & 0.352 & 0.386 & 0.342\\
    \cmidrule{2-12}
    & \multirow{7}{*}{merge}
    & Averaging & 0.520 & 0.581 & 0.547 & 0.572 & 0.214 & 0.567 & 0.522 & 0.489 & 0.502\\
    & & Fisher Merging & 0.533 & 0.594 & 0.564 & 0.579 & 0.224 & 0.577 & 0.529 & 0.500 & \bf0.512\\
    & & RegMean & 0.502 & 0.557 & 0.530 & 0.567 & 0.206 & 0.568 & 0.508 & 0.488 & 0.491\\
    & & Task Arithmetic & 0.337 & 0.353 & 0.347 & 0.295 & 0.110 & 0.330 & 0.274 & 0.278 & 0.291*\\
    & & TIES-Merging & 0.513 & 0.613 & 0.543 & 0.518 & 0.199 & 0.565 & 0.505 & 0.490 & 0.493\\
    & & AdaMerging & 0.474 & 0.149 & 0.200 & 0.156 & 0.162 & 0.360 & 0.407 & 0.365 & 0.284*\\
    & & PIM (Ours) & 0.438 & 0.581 & 0.671 & 0.668 & 0.186 & 0.480 & 0.407 & 0.434 & 0.483\\
    \cmidrule{2-12}
    & joint-train & - & 0.545 & 0.439 & 0.509 & 0.564 & 0.217 & 0.447 & 0.422 & 0.480 & 0.453*\\
    \bottomrule
    \end{tabular}
    \end{adjustbox}
    \caption{\textbf{Cross-dataset} cross-prompt evaluation results on \textbf{ASAP $\rightarrow$ PERSUADE2.0}, measured by QWK. For each target prompt in PERSUADE2.0, all $8$ prompts in ASAP constitute the source domains. }
    \label{tb:cross-dataset_result}
\end{table*}

\subsection{Experimental Setup}

\paragraph{Datasets.}
We validate our approach on two scenarios: (1) in-dataset cross-prompt scoring and (2) cross-dataset cross-prompt scoring. All samples are formatted using a simple instruction template, as in Appendix~\ref{apdx:instruction_template}.

\textbf{In-dataset cross-prompt scoring} follows the standard setup \citep{jin-etal-2018-tdnn, ridley2020prompt, li-ng-2024-conundrums} where each prompt in a dataset is held out as a target domain and the remaining prompts serve as source domains. We use the \textbf{ASAP}\footnote{\url{https://www.kaggle.com/c/asap-aes/data}} \citep{asap-aes} dataset, which includes essays written by students from grade $7$ to $10$ in response to $8$ prompts accross various genres and score ranges. We adopt the same dataset splits from \citet{ridley2021automated} where each prompt is split into training and validation sets approximately by $5.6:1$. When a prompt serves as the target domain, its two splits are combined into the test set. Dataset statistics are shown in Table~\ref{tab:dataset_stats}.

\textbf{Cross-dataset cross-prompt scoring} is a new setting we introduce to validate our approach and the baselines under severer distribution shifts. In particular, we use all prompts from ASAP as source domains and treat each prompt from \textbf{PERSUADE2.0} \citep{persuade2.0} as the target domain. PERSUADE2.0 contains essays written by U.S. students in response to $15$ prompts, among which we choose $4$ from independent writing and another $4$ from source-based writing during evaluation. Essay topics of the prompts are listed in Appendix~\ref{apdx:prompt_topics}.

\paragraph{Models \& Adapters.}
We conduct supervised fine-tuning on \textbf{Llama-3.1-8B-Instruct} \citep{grattafiori2024llama} ($8$ billion parameters) and \textbf{Phi-4-mini-instruct} ($4$ billion parameters) \citep{phi4}, for each prompt from the source domains independently. Details of LoRA fine-tuning are in Appendix \ref{apdx:training_details}. At inference, we use greedy decoding and parse the score from the model's response.

\paragraph{Baselines.}

We compare our approach against recent merging methods which we apply to source-free adaptation setting: \textbf{Averaging} \citep{wortsman2022model}, \textbf{Fisher Merging} \citep{matena2022merging}, \textbf{RegMean} \citep{jin2023dataless}, \textbf{Task Arithmetic} \citep{ilharco2023editing}, \textbf{TIES-Merging} \citep{yadav-ties} and \textbf{AdaMerging} \citep{yang2024adamerging}.

In addition, we report performance of \textbf{Joint-train} on all sources, and top-performing cross-prompt methods \textbf{PAES} \citep{ridley2020prompt} and \textbf{PMAES} \citep{chen-li-2023-pmaes}, all of which train the model jointly on all source domains. Particularly, PMAES requires unlabeled target samples simultaneously. See Appendix~\ref{apdx:baselines} for the descriptions and implementation details of the baselines.


\paragraph{Evaluation Metric.}
Following the standard evaluation protocol \citep{phandi2015Flexible, cao2020domain, chen-li-2023-pmaes}, we use the Quadratic Weighted Kappa (\textbf{QWK}) to measure the agreement between human-rated scores and predicted scores.

\paragraph{Implementation Details.}
We use Bayesian Optimization toolkit \citep{bayesopttoolkit}, with the Expected Improvement ($\xi=0.01$) acquisition function. We initially let the algorithm probe $10$ random points, and iterate through $30$ subsequent steps. Each coefficient $\lambda_j$ is bounded by $[0, 1]$. We randomly select $64$ test samples fixed across all iterations, and compute $p(y|x^{(i)}, \lambda)$ on those samples. If not otherwise stated, experimental results are averaged over $5$ random seeds.

\subsection{Main Results}
\paragraph{In-dataset Cross-prompt.} Results are shown in Table~\ref{tb:in-dataset_result}. First, our merging approach matches or surpasses joint training on all sources ($\textbf{0.602}$ vs. $0.601$ on Llama3.1-8B-it; $\textbf{0.629}$ vs. $0.587$ on Phi-4-mini-it), validating the effectiveness of selecting beneficial source domains for adaptation. In general, linear combination of task vectors underperforms its joint-training counterpart \citep{ilharco2023editing}, which makes the progress of our method over joint-training impressive. Second, our method exceeds all merging baselines in average QWK, with the improvements observed to be statistically significant, highlighting the importance of merging strategy specifically designed for domain adaptation. Third, our method brings a notable improvement over zero-shot baselines, with average gains of $0.301$ on Llama-3.1-8B-it and $0.362$ on Phi-4-mini-it. It also outperforms GPT-4.1-mini by $0.200$, showing our method's validity over a frontier LLM.

\paragraph{Cross-dataset Cross-prompt.} Table \ref{tb:cross-dataset_result} reports results on ASAP $\rightarrow$ PERSUADE2.0 transfer. Consistent with the in-dataset setting, our method shows improvements over joint training, e.g., by $0.111$ on Llama3.1-8B-it and $0.030$ on Phi-4-mini-it. Compared to the merging baselines, our method achieves the highest average QWK on Llama-3.1-8B-it and falls slightly short of some baselines on Phi-4-mini-it. Nevertheless, these baselines show limited generalizability, as they underperform on the other $3$ settings (Table \ref{tb:in-dataset_result}, \ref{tb:cross-dataset_result}) with larger drops, while our method maintains the best results. Overall, our approach shows robust adaptation on different types of domain shifts and LLMs. 

\begin{figure}[!t]
    \centering
    \includegraphics[width=\columnwidth]{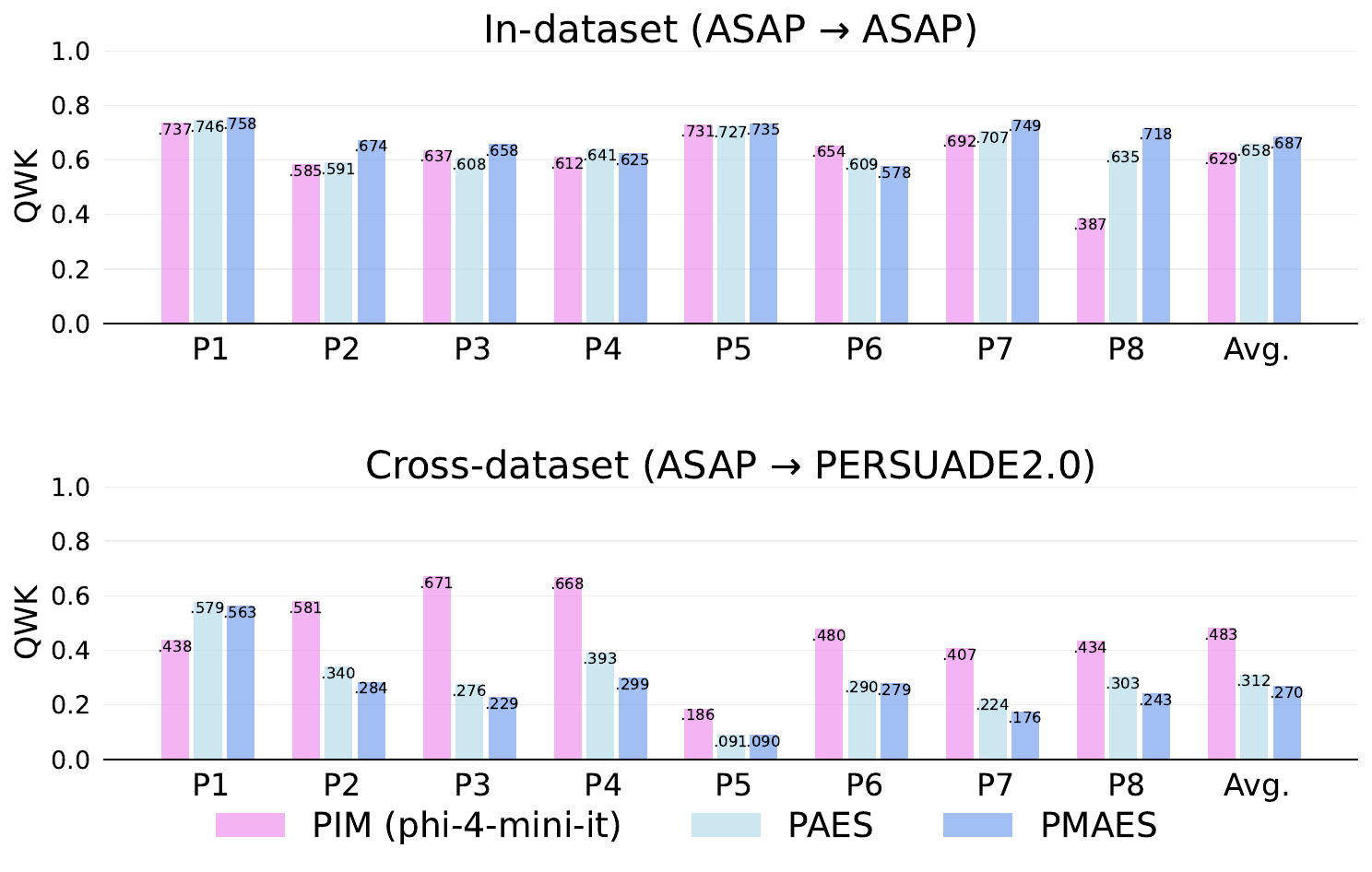}
    \caption{Comparison of PIM (phi-4-mini-it) with top-performing cross-prompt methods (PAES and PMAES). Similar trends for llama-3.1-8b-it (Appendix \ref{apdx:comparison_with_sota}).}
    \label{fig:comparison_with_sota_phi}
\end{figure}

\paragraph{Comparison with Leading Cross-prompt Methods.} Figure \ref{fig:comparison_with_sota_phi} compares our method with PAES \citep{ridley2020prompt} and PMAES \citep{chen-li-2023-pmaes} which show strong performance on \textbf{ASAP $\rightarrow$ ASAP}. Under in-dataset setting (\textbf{top}), our method achieves QWKs close to PAES and PMAES on most prompts but lags behind on average ($0.629$ vs. $0.658, 0.687$). In the more challenging cross-dataset setting (\textbf{bottom}), however, our method substantially outperforms PAES and PMAES with larger margins than in-dataset's case. This flip in superiority may in part be attributed to the two methods' reliance on all-source joint training and potentially domain-sensitive feature engineering. On the contrary, our method adaptively selects source domains to reduce negative transfer, 
which may have led to a balanced generalization across domain shifts with varying severity.

\begin{table}[t]
    \centering
    \small
    \resizebox{\columnwidth}{!}{
    \begin{tabular}{lcc}
        \toprule
        \textbf{Method} & \textbf{Phi4-mini} & \textbf{L3.1-8B} \\
        \midrule
        \textbf{PIM} & \bf0.629 & 0.602 \\ 
         ~$q(y)\rightarrow U$ & 0.594 & 0.590  \\
         ~w/o $-\mathcal{H}(p(y|x , \lambda))$ & 0.620 & \bf0.617  \\
         ~w/o $-KL(p(y|\lambda)||q(y))$  & 0.542 & 0.552  \\
         ~BayesOpt $\rightarrow$ Random & 0.611 & 0.595  \\
        \midrule
        joint-train & 0.587 & 0.601 \\
        \bottomrule
    \end{tabular}}
    \caption{Ablation study of PIM (Eq. \ref{eq:objective}), evaluated on ASAP $\rightarrow$ ASAP, measured by average QWK. w/o: without; BayesOpt $\rightarrow$ Random: same number of iterations with random search. }
    \label{tab:ablation}
\end{table}

\begin{figure}[!t]
    \centering
    \includegraphics[width=\columnwidth]{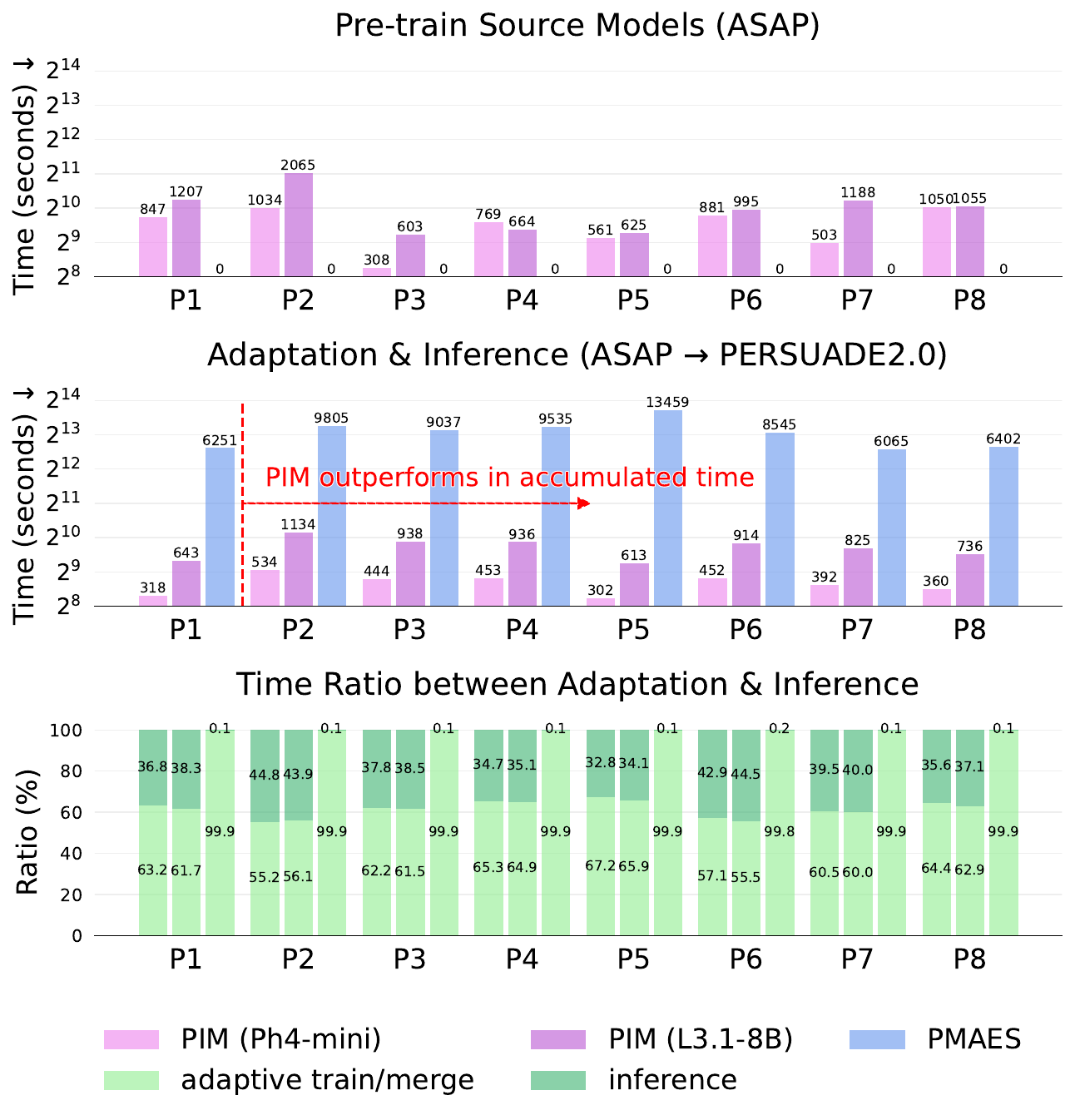}
    \caption{Log-scale time (y-axis) for pre-training source models on ASAP (\textbf{top}), followed by adaptation and inference on PERSUADE2.0 (\textbf{middle}), along with the time ratios between adaptation and inference (\textbf{bottom}). Results are from a single run on an NVIDIA A40 GPU.}
    \label{fig:time_cost_analysis}
\end{figure}

\subsection{Ablation Study}
We conduct an ablation study of the components of our method, as in Table \ref{tab:ablation}. \textbf{First,} reverting the useful prior back to the uniform distribution ($q(y)\rightarrow U$) leads to a notable drop in performance, which aligns with our motivation that $U$ may not appropriately represent the target's marginal distribution $p(y)$ in essay scoring. The prior $q(y)$ derived from the source prompts serves as a transferrable supervision signal, without any expert knowledge. \textbf{Second,} we challenge the MI maximization principle \citep{discluster10krause}, either by removing the sharpness term ($-\mathcal{H}(p(y|x , \lambda))$) or the separation term ($-KL(p(y|\lambda)||q(y))$). Interestingly, the former yields crossing result on the two models, with Llama-3.1-8B-it in fact achieving some gains. One hypothesis is that each source model (adapter) gives sufficiently sharp predictions on the target samples, and that merging the source models via linear combination preserves this property. In contrast, removing separation term leads to a significant drop, possibly due to its tendency to favor an overconfident model which lacks diversity in predictions. \textbf{Third,} given the same number of iterations, Bayesian optimization yields higher QWK than random search, verifying guided exploration.

\subsection{Cost Analysis}
In Figure~\ref{fig:time_cost_analysis}, we analyze the wall-clock time of PIM compared with PMAES, throughout the entire course of ASAP $\rightarrow$ PERSUADE2.0 adaptation. Notably, PIM requires substantially less adaptation$+$inference time than PMAES once all source models are trained. When accounting for the accumulated time from pre-training, PIM begins to outperform PMAES from the second target prompt, with the gap widening as more target prompts are introduced. This highlights PIM's scalability to new prompts, despite using larger models (LLMs) than PMAES (CNN+LSTM). The efficiency stems from the constant reuse of individual source models and efficient merging. PMAES retrains the entire model for each target, leading to high adaptation time relative to inference time.

\section{Related Work}

\paragraph{Cross-prompt Essay Scoring.}
Cross-prompt AES transfers models trained on source prompts to unseen ones \citep{dong-zhang-2016-automatic, ridley2021automated}. 
Early work used manual features and domain adaptation with a few labeled target samples \cite{phandi2015Flexible,cummins2016Constrained}.
Later neural methods improved generalization by incorporating prompt-agnostic objectives \citep{ridley2020prompt} or learning multi-prompt joint representations \cite{cao2020domain, chen-li-2023-pmaes}, but rely on static data fusion and full retraining, which limits the scalability. Our approach differs by enabling dynamic and selective utilization of source-domain knowledge without joint-training.

\paragraph{Model Merging.}

Model merging linearly combines parameters from same-architecture networks while preserving properties \cite{neyshabur2020being,zhou2023going}. 
Current methods include magnitude pruning \cite{yadav-ties,yu2024language,deep2024DELLAMerging,gargiulo2025Task,marczak2025No} to reduce parameter conflicts; activation merging \cite{yang2024representation,yang2024SurgeryV2,xu2025Scalable} to align features; optimization merging \cite{matena2022merging,jin2023dataless,yang2024adamerging} to adjust weights via optimization.
Emerging studies \cite{team2025Kimi,sun2025TinyR132BPreview} have demonstrated the effectiveness of model merging on LLMs, presenting a potential pathway for enhancing out-of-distribution performance.

\paragraph{Source-free Domain Adaptation.}

SFDA transfers models pretrained on labeled source domain(s) to unlabeled target domain without source data \citep{pmlr-v119-sun20b}. 
One approach includes compensating for the absence of source data by generating virtual samples \citep{tian22vdmda,ding22dist} or computing summary statistics \citep{adachi2025testtime}. Another approach is adapting solely with unlabeled target data by minimizing entropy \citep{wang2021tent,pmlr-v162-niu22a}, prompting prediction diversity \citep{pmlr-v119-liang20a,dong21confidence}, or Bayesian calibration \citep{zhou21bayesian}. Applications of SFDA in NLP are few, but growing \citep{zhang-etal-2021-matching,yin-etal-2024-source}. This paper addresses SFDA for essay scoring by leveraging source statistics, with model merging as a scalable alternative to training for adaptation.

\section{Conclusion}
In this paper, we propose a domain-adaptive model merging approach for source-free cross-prompt AES. Our pilot study suggests suboptimality of training on all source domains. Inspired by this, we shift to selecting beneficial source domains, and approximate costly joint training by merging task vectors through a linear combination. In optimizing the combination's coefficients, we resort to our proposed prior-encoded information maximization (PIM), an unsupervised objective which encourages score discriminability regularized by priors pre-computed from the sources. Experimental results with LLMs on in- and cross-dataset settings show that our method consistently outperforms joint training on all sources, surpasses other merging methods in numerous cases, maintains robustness under severe distribution shifts where leading cross-prompt methods struggle, all while remaining computationally efficient.

\section*{Limitations}

We elucidate the limitations of this work as follows: First, both mutual information maximization and our improved PIM rely on the assumption that at least one source model provides reasonable predictions (e.g., with sufficient diversity) for the target prompt. If all source models fail to capture the semantics of the target prompt, optimizing the PIM objective may degrade performance arbitrarily, as encouraging discriminability and sharpness becomes meaningless without meaningful initial predictions. Second, while our adaptation process remains efficient (using a fixed small sample of target essays, e.g., $64$), PIM is designed for LLMs, which incur significantly higher inference latency compared to conventional encoder-based AES models. This may limit scalability when each target prompt contains a very large volume of essays to be tested, despite the adaptation itself being sample-efficient. Third, although LLMs enable adaptation to novel score ranges, extreme deviations between source and target ranges can lead to suboptimal predictions. For instance, on  P8 of ASAP with score range of $[0, 60]$, some source models lacked diverse predictions.

\section*{Ethics Statement}
\paragraph{Potential Risks} This work aims to improve cross-prompt AES performance of LLMs. However, our method does not guarantee the model's fairness of scoring. For instance, it is possible that the adapted model assigns the scores in favor of a certain social group, such as the essay writer's first language background, gender, etc. In addition, since the source datasets may disproportionately represent certain social groups, models trained on these datasets could reproduce the biases embedded in the datasets in their predictions. There are ongoing works analyzing the fairness of AES systems \citep{loukina-etal-2019-many, schaller-etal-2024-fairness}, and it is recommended to refer to this field before the deployment of the system.

\paragraph{Use of Scientific Artifacts} For the datasets, we used ASAP \citep{asap-aes}'s publicly available text corpora, and used PERSUADE2.0 \citep{persuade2.0} which is an open source corpus under CC BY-NC-SA 4.0 license. Both ASAP and PERSUADE2.0 have anonymized personally identifying information from the essays. For the models, Llama-3.1-8B-Instruct \citep{grattafiori2024llama} is under Llama3.1 Community License, and Phi-4-mini-instruct \citep{phi4} is under MIT license. In addition, Bayesian Optimization \citep{bayesopttoolkit} toolkit is under MIT license. All of these artifacts is applicable for research use.


\bibliography{custom}

\appendix

\section{Details of Pilot Study}
\label{apdx:pilot_study_full}
\begin{figure*}[tbp]
    \centering
    \includegraphics[width=1.0\textwidth]{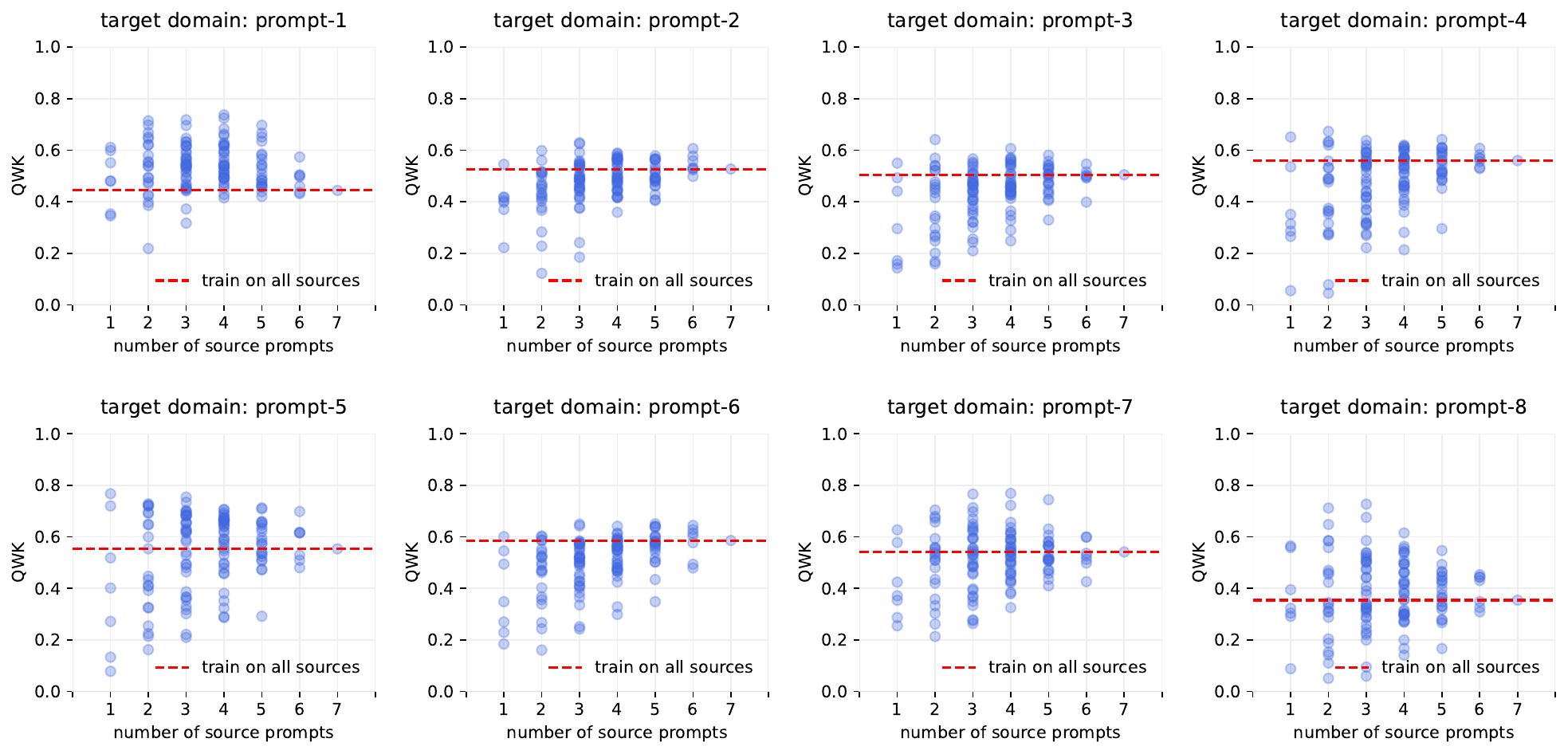}
    \caption{ Agreement with human raters (QWK) on target domains using BERT \citep{devlin-etal-2019-bert} trained jointly on varying number of source prompt datasets. }
    \label{fig:pilot_study_full}
\end{figure*}

We investigate the effect of jointly training on varying number of source datasets on the transfer performance. In this experiment, we train BERT with the following configurations: a regression module is placed on top of BERT, which consists of a linear layer and a sigmoid layer; We train $30$ epochs with a batch size of $128$, a learning rate of $2\cdot10^{-5}$ with constant learning schedule; We choose the best checkpoint on the validation set among the epochs, with early stopping of $10$ epochs. Figure~\ref{fig:pilot_study_full} shows the transfer performance of training on all possible subsets of the source prompt datasets. We observe that regardless of the target prompt, training on all sources is suboptimal, and a careful selection of the sources improves transfer.

\section{Supplementary Details of Method}

\subsection{Matching Mean and Variance of Beta Mixture}
\label{apdx:moment_matching}
We describe how the unified $\mathbf{Beta}(\alpha_{\mathcal{S}}, \beta_{\mathcal{S}})$ is derived by matching its mean and variance to the Beta mixture $1/M\sum_{j=1}^{M} \mathbf{Beta}(\alpha_j, \beta_j)$. For each source distribution $\mathbf{Beta}(\alpha_j, \beta_j)$, the mean $\mu_j$ and variance $\sigma^2_j$ are
\begin{equation*}
    \mu_j = \frac{\alpha_j}{\alpha_j + \beta_j}, \sigma^2_j =\frac{\alpha_j\beta_j}{(\alpha_j+\beta_j)^2(\alpha_j+\beta_j+1)}.
\end{equation*}
The mean $\mu$ and variance $\sigma^2$ of the mixture is then given by
\begin{align*}
    \mu &= M^{-1}\sum_{j=1}^M \mu_j \\
    \sigma^2 &= M^{-1}\sum_{j=1}^M (\sigma^2_j + \mu^2_j) - \mu^2.
\end{align*}
Finally, we let $\mathbf{Beta}(\alpha_{\mathcal{S}}, \beta_{\mathcal{S}})$ have $\mu$ and $\sigma^2$ as its mean and variance, in essence:
\begin{align*}
    \alpha_{\mathcal{S}} &:= \mu\left( \frac{\mu(1-\mu)}{\sigma^2} -1 \right) \\
    \beta_{\mathcal{S}} &:= (1-\mu)\left( \frac{\mu(1-\mu)}{\sigma^2} -1 \right).
\end{align*}

\subsection{Details of Bayesian Optimization}
\label{apdx:bayes_opt_details}

Bayesian optimization \citep{williams2006gaussian} constructs a surrogate model of the black-box function $f(\lambda)$ as a sample from a Gaussian Process---a distribution over functions, and updates the posterior on $f$ given observations $\{f(\lambda^{(i)})\}_{i=1}^{k}$. It then uses an \textit{acquisition function} to determine where to sample $\lambda^{(k+1)}$ next. When the iteration terminates,  $\lambda^*=\mathop{\arg \max}_{\lambda^{(i)}} f(\lambda^{(i)})$ is chosen as the final solution. 

In detail, for $(k+1)$-iteration, the Gaussian prior is placed on the observations:
\begin{equation*}
    f(\lambda^{(1:k)}) \sim \mathcal{N}(\mu_{0}(\lambda^{(1:k)}), \Sigma_{0}(\lambda^{(1:k)}, \lambda^{(1:k)})) 
\end{equation*}
where $\lambda^{(1:k)}$ is a compact notation for $k$ points $\{f(\lambda^{(i)})\}_{i=1}^{k}$, and $\mu_{0}$ and $\Sigma_{0}$ are the mean and covariance function of the Gaussian Process. We choose the commonly used $\mathbf{0}$ for $\mu_0$ and Matern 2.5 kernel \citep{williams2006gaussian} for $\Sigma_0$. Then the posterior on a new function value $f(\lambda^{(k+1)})$ given previous observations $f(\lambda^{(1:k)})$ is updated by the Bayes' rule \citep{frazier2018tutorial}:
\begin{multline*}
f(\lambda^{(k+1)})|f(\lambda^{(1:k)})  \\ \sim \mathcal{N}(\mu_{k+1}(\lambda^{(k+1)}), \sigma^2_{k+1}(\lambda^{(k+1)}))
\end{multline*}
where
\begin{multline*}
    \mu_{k+1}(\lambda^{(k+1)})=\Sigma_{0}(\lambda^{(k+1)}, \lambda^{(1:k)}) \\ \cdot \Sigma_{0}(\lambda^{(1:k)}, \lambda^{(1:k)})^{-1}\\
    \cdot(f(\lambda^{(1:k)})- \mu_{0}(\lambda^{(1:k)})) + \mu_{0}(\lambda^{(k+1)})
\end{multline*}
\begin{multline*}
    \sigma^2_{k+1}(\lambda^{(k+1)})=\Sigma_{0}(\lambda^{(k+1)}, \lambda^{(k+1)}) \\ - \Sigma_{0}(\lambda^{(k+1)}, \lambda^{(1:k)})\\ \cdot\Sigma_{0}(\lambda^{(1:k)}, \lambda^{(1:k)})^{-1}\Sigma_{0}(\lambda^{(1:k)},\lambda^{(k+1)}).
\end{multline*}
Intuitively, the posterior mean $\mu_{k+1}(\lambda^{(k+1)})$ is a weighted sum between the prior $\mu_0(\lambda^{(k+1)})$ and a calibration term based on the data $f(\lambda^{(1:k)})$, and the posterior variance $\sigma^2_{k+1}(\lambda^{(k+1)})$ is given as the prior variance $\Sigma_{0}(\lambda^{(k+1)}, \lambda^{(k+1)})$ subtracted by the reduction in variance (uncertainty) after observing the data $f(\lambda^{(1:k)})$ \citep{frazier2018tutorial}.

Next, the acquisition function specifies where to sample $\lambda^{(k+1)}$ based on the posterior. We use Expected Improvement (\textbf{EI}) which finds $\lambda^{(k+1)}$ such that the expected gain over the current best value $f^*(k):=\mathop{\max}_{\lambda^{(i)}}\{f(\lambda^{(i)})\}_{i=1}^{k}$ is maximized:
\begin{equation*}
    \mathop{\arg \max}_{\lambda^{(k+1)}} \mathop{\mathbb{E}}_{f(\lambda^{(k+1)})}\left[ \max(f(\lambda^{(k+1)}) - f^*(k), 0) \right]
\end{equation*}
This process of posterior estimation and next point sampling is repeated until convergence.

\section{Additional Details of Experimental Setup}
\subsection{Instruction Template}
\label{apdx:instruction_template}
We use the instruction template below throughout the experiments.

\noindent
{\ttfamily
    \textbf{\underline{User Message}}\\
    \#\#\# Prompt:\\
    \{prompt\}\\
    \#\#\# Student Essay:\\
    \{essay\}\\
    \#\#\# Instruction:\\
    Given the student's essay written in response to the prompt, assign a score within the range of \{min\_score\} to \{max\_score\}. Respond with only an integer score and no additional text.\\
}
{\ttfamily
    \textbf{\underline{Assistant Message}}\\
    \{score\}\\
}

\subsection{Prompt Topics}
\label{apdx:prompt_topics}
In Table \ref{tab:prompt_topics}, we specify the correspondence between the prompt IDs (Table. \ref{tab:dataset_stats}) and the prompt topics.

\begin{table}[ht]
    \centering
    \small
    \resizebox{\columnwidth}{!}{
    \begin{tabular}{ccl}
        \toprule
        \textbf{Dataset} & \textbf{Prompt} & \textbf{Topic} \\
        \midrule
        \multirow{8}{*}{\textbf{ASAP}}
         & 1 & Effects computers have on people  \\
         & 2 & Censorship in the libraries  \\
         & 3 & Impact of setting on the cyclist’s experience  \\
         & 4 & The meaning of the ending in Winter Hibiscus  \\
         & 5 & The mood created in Narciso Rodriguez’s memoir  \\
         & 6 & Obstacles to docking dirigibles \\
         & 7 & A story about patience  \\
         & 8 & A story about laughter \\
        \midrule
        \multirow{8}{*}{\textbf{PERSUADE2.0}}
        & 1 & Cell phones at school \\
         & 2 & Distance learning \\
         & 3 & Mandatory extracurricular activities \\
         & 4 & Seeking multiple opinions \\
         & 5 & "A Cowboy Who Rode the Waves" \\
         & 6 & Does the electoral college work? \\
         & 7 & Exploring Venus \\
         & 8 & The Face on Mars \\
        \bottomrule
    \end{tabular}}
    \caption{Prompt Topics of ASAP and PERSUADE2.0.}
    \label{tab:prompt_topics}
\end{table}

\subsection{Details of LoRA Fine-tuning}
\label{apdx:training_details}
We use LoRA with $r=16$, $\alpha=32$, $\text{dropout}=0.1$, targeting all linear layers in the transformer block \citep{transformer}. During fine-tuning we use the AdamW optimizer \citep{loshchilov2018decoupled} with a batch size of $16$, a learning rate of $10^{-4}$ and a cosine scheduler. The best checkpoint on the validation set is selected, with evaluation steps of $30$ and early stopping patience of $3$. 

\subsection{Descriptions and Implementation Details of Baselines}
\label{apdx:baselines}

\textbf{Averaging} \citep{wortsman2022model} simply averages the models' parameters. \textbf{Task Arithmetic} \citep{ilharco2023editing} adds a scaled sum of task vectors to the pre-trained model, and \textbf{TIES-Merging} \citep{yadav-ties} pre-processes task vectors to resolve their interferences prior to merging. Following the recommended hyperparameters, we set the scaling factor of TA to $\lambda=0.4$ and of TIES to $\lambda=1.0$. 

\textbf{Fisher Merging} \citep{matena2022merging} improves Averaging by accounting for parameter-wise importance using Fisher information. Fisher information is estimated by sampling from the label distribution of samples from the validation set. \textbf{RegMean} \citep{jin2023dataless} aims to minimize the layer-wise distance in activation between the merged model and all fine-tuned models. Following the original implementation on T5 models, we set the non-diagonal multiplier to $\alpha=0.1$. 

\textbf{AdaMerging} \citep{yang2024adamerging} is a test time adaptation method which trains layer-wise coefficients for merging task vectors in order to minimize entropy on test samples. In our domain adaptation setting, test samples are the samples from the target domain. In contrast to other baselines, AdaMerging exploits the information of the target domain samples. 

As for \textbf{Joint-train} baseline, we use the same configuration as \ref{apdx:training_details} except for $\text{batch size}=64$ and $\text{early stopping patience}=10$. For another joint-training baselines, \textbf{PAES} \citep{ridley2020prompt} and \textbf{PMAES} \citep{chen-li-2023-pmaes}, we follow the original settings for model architecture and training hyperparameters. As the original implementation of PMAES does not specify the batch size, we set the combined batch size for the source and target domains data to 32, and allocate it proportionally based on the data ratio between the source and target domains.

\section{Additional Comparison with Leading Cross-prompt Methods}
\label{apdx:comparison_with_sota}
\begin{figure}[t]
    \centering
    \includegraphics[width=\columnwidth]{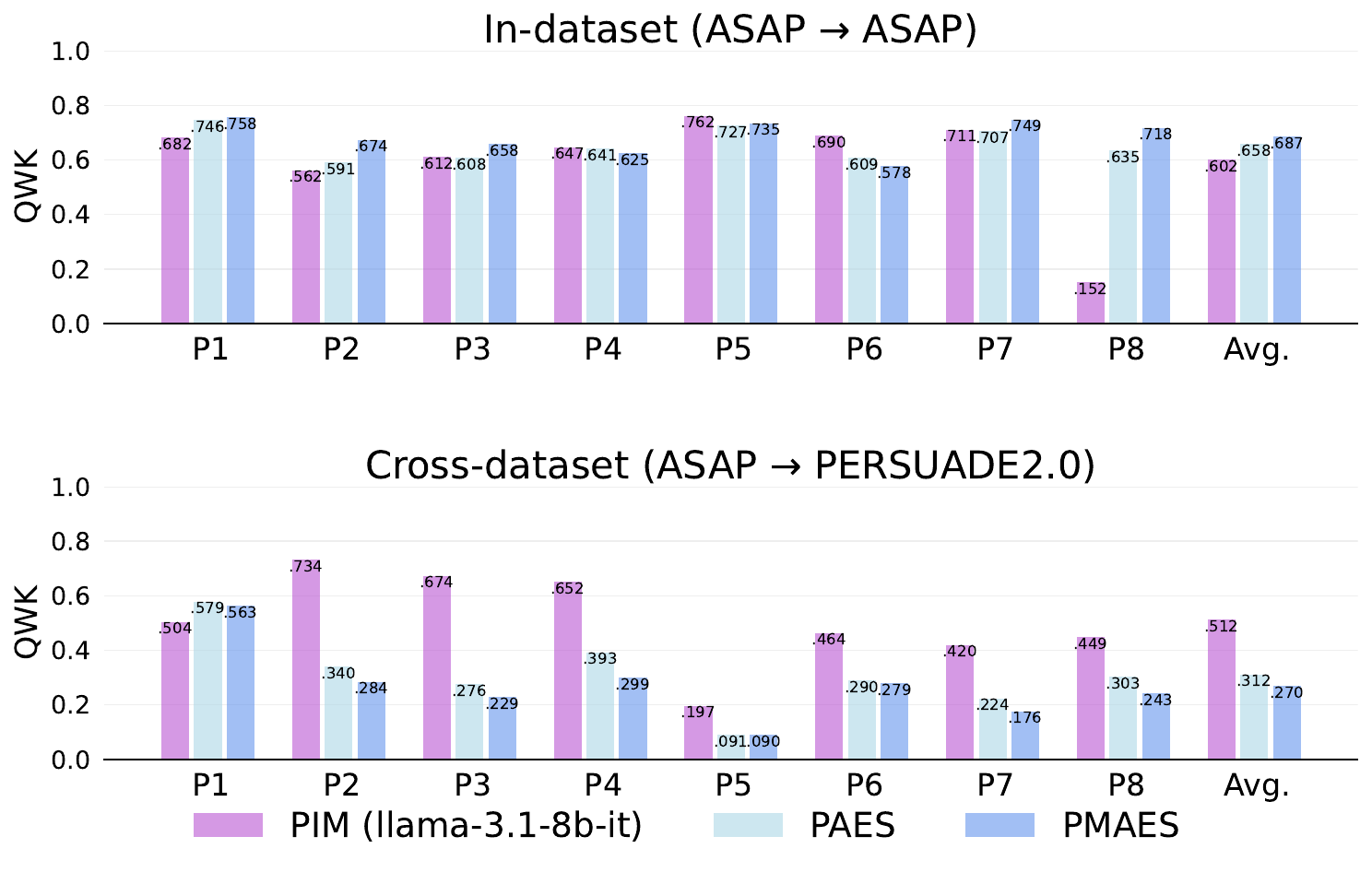}
    \caption{Comparison of PIM (llama-3.1-8b-it) with top-performing cross-prompt methods (PAES and PMAES).}
    \label{fig:comparison_with_sota_llama}
\end{figure}

In Figure \ref{fig:comparison_with_sota_llama}, we show additional comparison results between PIM (llama-3.1-8b-it) and leading cross-prompt AES methods---PAES \citep{ridley2020prompt} and PMAES \citep{chen-li-2023-pmaes}. The overall trend is consistent with  PIM (phi-4-mini-it).

\end{document}